\documentclass[conference]{IEEEtran}
\IEEEoverridecommandlockouts

\usepackage{cite}
\usepackage{amsmath,amssymb,amsfonts}
\usepackage{algorithmic}
\usepackage{graphicx}
\usepackage{textcomp}
\usepackage{xcolor}
\usepackage{hyperref}
\usepackage{booktabs}
\usepackage{multirow}
\usepackage{stfloats}  
\usepackage{afterpage} 

\graphicspath{{./figures/media/}}

\title{Reward-Density Heuristic for Dynamic Multi-Vehicle Routing: Performance and Computational Efficiency}

\author{
  \IEEEauthorblockN{Manish Kolachalam}
  \IEEEauthorblockA{\textit{Autonomous Machines Applied Research Center} \\
  \textit{Infosys Center for Imaging Technologies, Infosys} \\
  Bengaluru, India \\
  manish.k02@infosys.com}
  \and
  \IEEEauthorblockN{Rani Malhotra}
  \IEEEauthorblockA{\textit{Autonomous Machines Applied Research Center} \\
  \textit{Infosys Center for Imaging Technologies, Infosys} \\
  Bengaluru, India \\
  rani.malhotra@infosys.com}
}

\begin{document}

\maketitle

\begin{abstract}
The Vehicle Routing Problem (VRP) and its variants represent some of the most practically consequential optimization challenges in modern logistics and urban mobility. We address a dynamic, online variant combining elements of the VRP and the Orienteering Problem (OP), in which a fleet of vehicles must maximise cumulative reward within a fixed time horizon while continuously replanning as new tasks arrive. We propose and evaluate a reward-density heuristic for dynamic multi-vehicle assignment (the Efficiency heuristic) across two domains --- autonomous drone task allocation and urban taxi dispatch --- at multiple fleet sizes. The method is compared with four construction heuristics and three metaheuristic algorithms (ALNS, GA, SA) under identical conditions. Across all tested configurations, the Efficiency heuristic matches the solution quality of the best metaheuristics while requiring two to three orders of magnitude less planning time, establishing Pareto dominance over all competing methods on the reward-versus-compute frontier.
\end{abstract}


\section{Introduction}

In 2018, U.S. businesses spent 10.4\% of their revenue on transportation costs alone, while overall logistics expenditure constituted 8\% of GDP \cite{b1}. The Vehicle Routing Problem (VRP), first formalized by Dantzig and Ramser in 1959 \cite{b2}, is one of the most extensively studied problems in combinatorial optimization. The VRP and the Travelling Salesman Problem (TSP) \cite{b3} are both NP-hard, making exact solutions intractable at scale \cite{b4}, with broad implications for fuel consumption, delivery times, and carbon emissions \cite{b5}.

A particularly challenging variant is the dynamic VRP, in which tasks arrive online and vehicles must replan continuously \cite{b6,b7}. We focus on a formulation combining the dynamic VRP with the Orienteering Problem (OP) \cite{b8}, where vehicles maximise cumulative reward within a fixed time horizon. This structure arises in drone task allocation \cite{b9,b10} and urban taxi dispatch \cite{b11,b12}: in both settings, tasks arrive dynamically, vehicles are allocated based on current location, and the binding constraint is a time budget. This aligns our formulation with the prize-collecting VRP \cite{b13} and the OP \cite{b8}, where selective task completion is central, and both domains require assignment decisions within milliseconds.

A natural candidate for such settings is a greedy reward-density heuristic --- scoring each candidate task by reward divided by time cost \cite{b14}. Such a rule requires no population, no iteration, and no convergence. Whether this simplicity comes at a meaningful cost in solution quality relative to metaheuristics is an open empirical question with direct practical implications.

This paper evaluates that question across five experimental configurations. The Efficiency heuristic consistently matches ALNS, GA, and SA while requiring two to three orders of magnitude less planning time, across synthetic drone environments and real-world NYC taxi dispatch data \cite{b15}. Contributions: (1) a unified dynamic reward-maximising VRP framework; (2) two reward-density instantiations --- greedy sequential and optimal Hungarian matching \cite{b16}; and (3) systematic demonstration of Pareto-optimal performance across five configurations.

\section{Related Work}

\subsection{Construction Heuristics}

\textbf{Nearest Neighbour (Greedy-Nearest):} The nearest-neighbour heuristic \cite{b20} greedily selects the closest unvisited node, producing routes in polynomial time and serving as a standard VRP baseline \cite{b21}.

\textbf{Reward-Greedy (Greedy-Reward):} Reward-greedy heuristics prioritise high-value targets regardless of spatial location, and are natural choices for prize-collecting and orienteering formulations \cite{b22}. They have been widely applied in UAV task allocation \cite{b9}, though they may commit vehicles to distant targets while neglecting nearby ones of comparable value.

\textbf{Hungarian Algorithm:} The Hungarian algorithm solves the LAP optimally in polynomial time \cite{b16} and has been widely applied in multi-robot task allocation and online dispatch \cite{b23}. We evaluate two variants: minimising travel time (Hungarian-Time) and maximising raw reward (Hungarian-Reward) using the SciPy \cite{b18} implementation.

\subsection{Metaheuristic Algorithms}

Genetic algorithms \cite{b24} evolve candidate solutions through selection and mutation, producing near-optimal solutions on static instances given sufficient time. Simulated annealing \cite{b25} escapes local optima via a temperature-dependent acceptance criterion at lower cost than population-based methods, making it a candidate for time-constrained settings. ALNS \cite{b26} maintains a portfolio of destroy and repair operators selected adaptively, with strong performance across VRP variants. Because the assignment landscape changes with every task completion, all metaheuristic planners are reinitialized at each replanning event.

\subsection{Efficiency Heuristics}

The reward-density principle is rooted in the OP, where Tsiligirides \cite{b14} introduced a desirability measure $A(j) = S_j / t_{\text{last},j}$ (reward per unit travel time) as the basis for the S- and D-algorithms. The structure also resembles the fractional knapsack relaxation \cite{b27} --- density-greedy is optimal for the fractional case --- and has appeared in fractional programming \cite{b28} and time-constrained orienteering \cite{b29}. For UAV operations, \cite{b30} proposed a greedy best-aisle heuristic selecting targets by maximum reward-to-cost ratio, directly analogous to our scoring rule. In dynamic settings, \cite{b6} established that static solutions degrade under online arrivals, motivating fast reactive replanning. The present work extends this to a fully dynamic, multi-vehicle setting with systematic metaheuristic comparison across two domains.

\section{Methods}

\begin{table*}[htbp]
\centering
\caption{Comparison of drone dispatch algorithms in terms of reward and planning time for fleets of 12, 20, and 50 drones. Reference algorithm is Hungarian-Efficiency; statistical significance relative to the reference is indicated by \textsuperscript{ns} (not significant), \textsuperscript{*} ($p < 0.05$), \textsuperscript{**} ($p < 0.01$), or \textsuperscript{***} ($p < 0.001$).}
\label{tab:drone_dispatch_comparison}
\begin{tabular}{l cc cc cc}
\toprule
\multirow{2}{*}{\textbf{Algorithm}} & \multicolumn{2}{c}{\textbf{12 Drones}} & \multicolumn{2}{c}{\textbf{20 Drones}} & \multicolumn{2}{c}{\textbf{50 Drones}} \\
\cmidrule(lr){2-3} \cmidrule(lr){4-5} \cmidrule(lr){6-7}
 & Reward & Plan time (ms) & Reward & Plan time (ms) & Reward & Plan time (ms) \\
\midrule
Greedy-Nearest        & $1{,}556 \pm 63^{**}$   & $1.66 \pm 0.79$    & $3{,}048 \pm 109^{*}$   & $17.29 \pm 1.04$   & $7{,}098 \pm 174^{***}$ & $71.43 \pm 2.16$   \\
Greedy-Reward         & $1{,}636 \pm 48^{*}$    & $6.32 \pm 1.72$    & $3{,}004 \pm 50^{***}$  & $12.54 \pm 1.85$   & $7{,}266 \pm 107^{***}$ & $68.06 \pm 2.15$   \\
Greedy-Efficiency     & $1{,}850 \pm 57^{ns}$   & $7.08 \pm 1.90$    & $3{,}585 \pm 85^{ns}$   & $22.70 \pm 1.79$   & $8{,}308 \pm 162^{ns}$  & $104.06 \pm 3.70$  \\
Hungarian-Time        & $1{,}635 \pm 67^{*}$    & $3.42 \pm 1.24$    & $3{,}078 \pm 88^{**}$   & $20.56 \pm 1.57$   & $7{,}275 \pm 166^{***}$ & $92.67 \pm 2.78$   \\
Hungarian-Reward      & $1{,}595 \pm 46^{*}$    & $7.33 \pm 1.81$    & $3{,}003 \pm 62^{***}$  & $16.96 \pm 0.57$   & $7{,}228 \pm 121^{***}$ & $84.73 \pm 2.98$   \\
Hungarian-Efficiency  & $1{,}811 \pm 53$ (ref)  & $4.33 \pm 1.51$    & $3{,}485 \pm 91$ (ref)  & $25.93 \pm 1.86$   & $8{,}306 \pm 175$ (ref) & $127.92 \pm 3.18$  \\
\midrule
ALNS                  & $1{,}797 \pm 62^{ns}$   & $1{,}908 \pm 79$   & $3{,}502 \pm 90^{ns}$   & $6{,}737 \pm 229$  & $8{,}430 \pm 161^{ns}$  & $42{,}654 \pm 3{,}046$ \\
GA                    & $1{,}809 \pm 50^{ns}$   & $15{,}790 \pm 228$ & $3{,}530 \pm 87^{ns}$   & $50{,}848 \pm 541$ & $8{,}482 \pm 175^{ns}$  & $1.10M \pm 0.85M$  \\
SA                    & $1{,}812 \pm 59^{ns}$   & $334.58 \pm 4.28$  & $3{,}542 \pm 82^{ns}$   & $1{,}060 \pm 12$   & $8{,}419 \pm 180^{ns}$  & $5{,}247 \pm 101$  \\
\bottomrule
\end{tabular}
\end{table*}

\subsection{Problem Formulation and Simulation}

A fleet of $V$ vehicles operates from a central depot. A set of $N$ targets is distributed across an $M \times M$ map, each target $i$ with location $(x_i, y_i)$, reward $r_i \sim \mathcal{U}(1,100)$, and service time $s_i \sim \mathcal{U}[5,30]$ time units. Vehicle speeds are drawn from $\{10, 15, 30\}$ map units/time unit. A vehicle may only be assigned target $i$ if travel time plus service time does not exceed its remaining budget. The objective is to maximise total reward across all vehicles within the mission duration.

We implement an event-driven simulator where replanning triggers on task completion. Only vehicles that just completed a task receive new assignments; en-route vehicles are not interrupted. Simultaneous completions are batched into a single replanning call. A shared claimed set prevents duplicate assignments. Performance is evaluated over 15 independent runs (drone) and 10 runs (taxi) on identical environments from shared random seeds.

\subsection{Experimental Configuration}

\textbf{Drone:} Three configurations --- 12/50, 20/100, 50/200 vehicles/targets --- over 200 time-units on a $5000\times5000$ map, with rewards drawn uniformly from 1--100. \textbf{Taxi:} Fleet sizes of 50 and 200 vehicles over 120 minutes against $\sim$7,600 tasks from the NYC TLC Yellow Cab 2024 dataset \cite{b15}. Vehicles follow passenger trips sequentially from the drop-off location; revenue uses actual TLC fare data. This dataset provides realistic spatial clustering, heterogeneous trip lengths, and empirical arrival patterns characteristic of stochastic VRP settings \cite{b13,b31}.

\begin{figure*}[htbp]
\centering
\includegraphics[width=\textwidth]{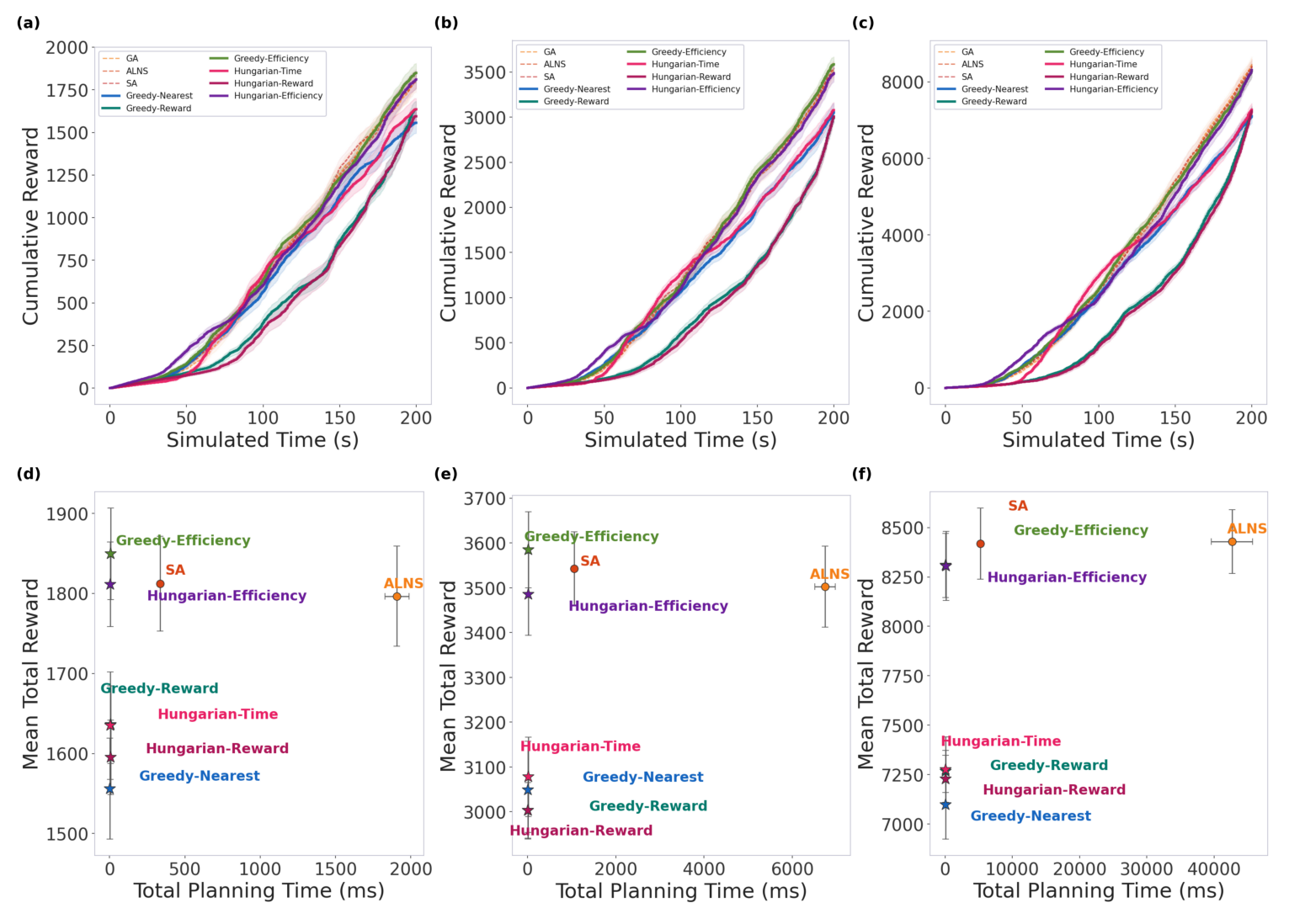}
\caption{Performance comparison of multi-drone task allocation algorithms across three fleet sizes. (a--c) Cumulative reward over simulated time (0--200 s) for increasing numbers of drones, showing mean trajectories with shaded confidence intervals. (d--f) Mean total reward versus total planning time (ms), illustrating the reward-efficiency trade-off (GA omitted for scale).}
\label{fig:drone}
\end{figure*}

\subsection{Algorithms}

\textbf{Construction heuristics} share a greedy framework: feasible targets are scored and each free vehicle takes the highest-scoring unclaimed target, with vehicles processed in index order. \textit{Greedy-Nearest} scores by negative distance; \textit{Greedy-Reward} by reward value; \textit{Hungarian-Time} solves a LAP minimising travel times via SciPy \cite{b18}; \textit{Hungarian-Reward} solves the same LAP maximising reward.

\textbf{Metaheuristics} (GA, SA, ALNS) encode the assignment as optimisation over a length-$V$ priority vector. GA uses epoch\,=\,100, population\,=\,50 (MealPy \cite{b17}). SA uses population\,=\,2, cooling factor 0.95, epoch\,=\,100, with initial temperature calibrated to 15\% of mean target reward. ALNS \cite{b19} uses 200 iterations/step (drone) or 100 (taxi); destroy operators are random removal and worst-50\% removal by reward-to-time ratio; repair operators are greedy-efficiency and regret-based; roulette selection uses scores $[5,3,1,0]$ with decay 0.8, warm-started with a greedy-efficiency solution.

\textbf{Reward-density heuristics.} Each task consumes travel time plus service time from a vehicle's fixed budget, analogous to the fractional knapsack. The desirability score for candidate target $\tau$ is:
\begin{equation}
\eta(\tau) = \frac{R(\tau)}{d(v,\tau) + s(\tau)}
\end{equation}
where $R(\tau)$ is reward, $d(v,\tau)$ is travel time from vehicle position $v$, and $s(\tau)$ is service time. This extends Tsiligirides \cite{b14} to a dynamic multi-vehicle setting by incorporating service time. \textit{Greedy-Efficiency} assigns each free vehicle its highest-scoring feasible unclaimed target sequentially. \textit{Hungarian-Efficiency} solves a LAP using $-\eta(\tau)$ as the cost matrix, finding the globally optimal assignment under the reward-density objective \cite{b16}.

\subsection{Statistical Analysis}

We report mean $\pm$ SEM across runs. Statistical significance uses Bonferroni-corrected Mann-Whitney U tests against Hungarian-Efficiency: $^*p{<}0.05$, $^{**}p{<}0.01$, $^{***}p{<}0.001$; ns\,=\,not significant. Cumulative reward curves are averaged by interpolating each run's step function onto a 500-point time grid.

\section{Results}

We evaluated four heuristic approaches (Greedy-Nearest, Greedy-Reward, Hungarian-Time and Hungarian-Reward) and three metaheuristic planners (GA, SA and ALNS) across two dynamic resource allocation problems: online drone task assignment and taxi dispatch. Performance was measured using cumulative reward (or revenue) and planning cost. All results represent averages over 15 independent runs for the drone experiments and 10 runs for the taxi experiments, with error bars corresponding to the standard error of the mean (SEM) and boxplots showing interquartile ranges as well as the mean$\pm$SEM. The drone simulations were run for 3 different set of task parameters and the taxi dispatch experiments for two task parameters.

\subsection{Drone Dispatch}

Across all fleet sizes tested (12, 20, and 50 drones), the efficiency-optimised algorithms consistently outperformed simpler heuristics on revenue while remaining computationally practical. Hungarian-Efficiency served as the primary reference, achieving reward of 1,811, 3,485, and 8,306 (mean) at 12, 20, and 50 drones respectively. The Greedy-Efficiency algorithm matched this performance closely at all scales.

The simpler greedy approaches --- Greedy-Nearest and Greedy-Reward --- both fell significantly short of Hungarian-Efficiency at every fleet size ($p < 0.05$ to $p < 0.001$) and Hungarian-Time and Hungarian-Reward similarly underperformed, achieving revenues significantly below the Efficiency heuristics at all scales ($p < 0.001$).

The metaheuristic approaches --- ALNS, GA, and SA --- all garnered reward statistically comparable to the Efficiency heuristics (ns at all fleet sizes), but at dramatically higher planning costs. At 50 drones, SA required $\sim$5,247 ms, ALNS $\sim$42,654 ms, and GA over 1.1 million ms, compared to $\sim$128 ms for Hungarian-Efficiency. This represents planning overhead of roughly $40\times$, $330\times$, and $8{,}600\times$ respectively, making these methods impractical for real-time dispatch at scale.

Results in the taxi domain largely replicated the drone findings, with efficiency-based methods dominating on the revenue-versus-speed trade-off. At both 50 and 200 taxis, Hungarian-Efficiency and Greedy-Efficiency achieved statistically equivalent revenues (ns) and significantly higher than the other heuristics ($p < 0.001$). The three metaheuristics (ALNS, GA, SA) produced revenues statistically indistinguishable from Hungarian-Efficiency and Greedy-Efficiency, but incurred prohibitive planning times. At 200 taxis, GA required $\sim$4.58 million ms per planning cycle, SA $\sim$96,660 ms, and ALNS $\sim$236,765 ms --- versus 363 ms for Hungarian-Efficiency. These costs render all three impractical for operational dispatch.

\begin{figure*}[htbp]
\centering
\includegraphics[width=0.8\textwidth]{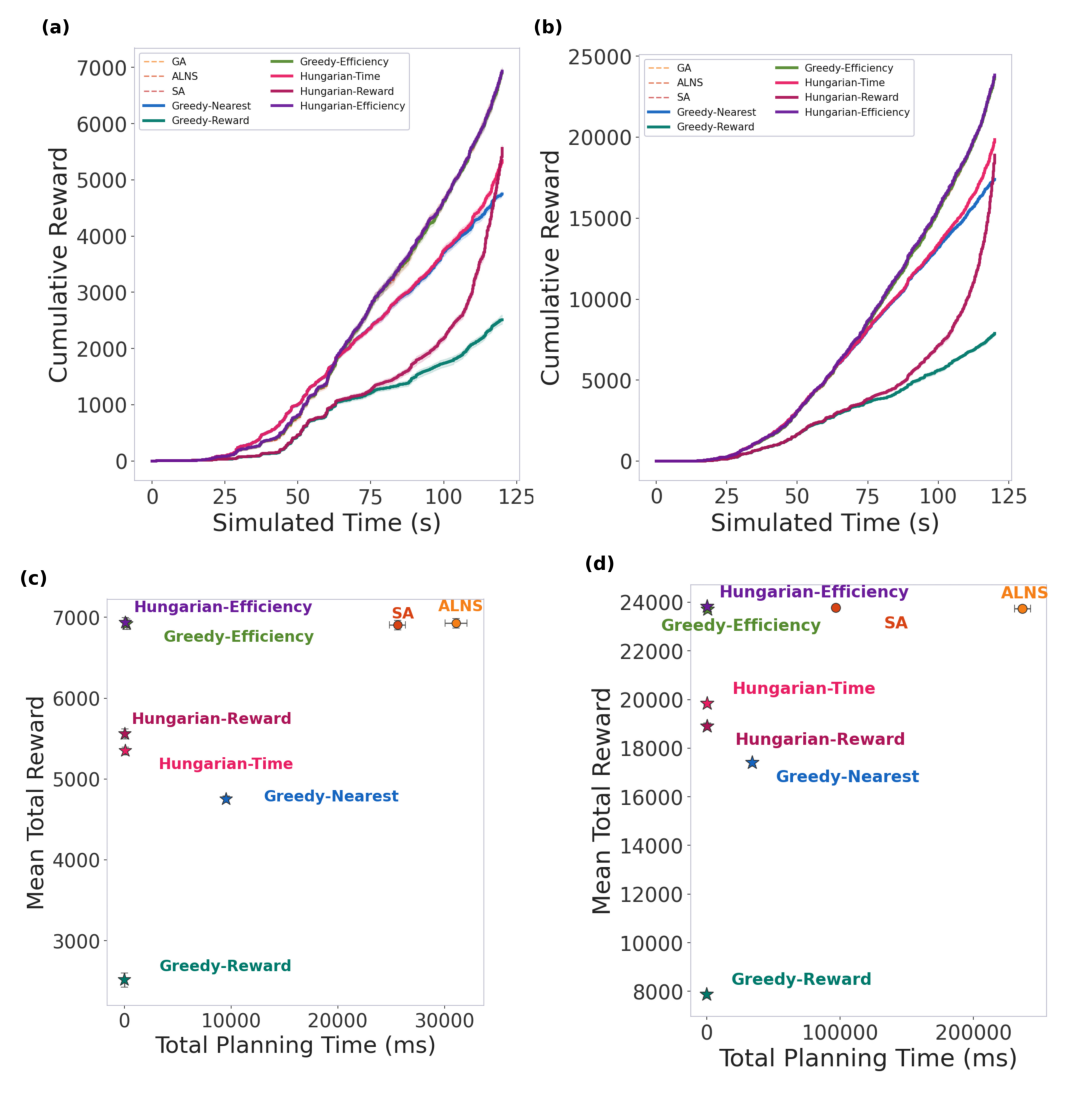}
\caption{Performance comparison of taxi dispatch algorithms across two fleet sizes. (a--b) Cumulative reward over simulated time. (c–d) Mean total reward versus total planning time (ms). Error bars denote standard deviation across trials (GA omitted for scale).}
\label{fig:drone}
\end{figure*}

\begin{table*}[htbp]
\centering
\caption{Comparison of taxi dispatch algorithms in terms of revenue and planning time across fleet sizes of 50 and 200 taxis. Values are reported as mean $\pm$ standard deviation. Statistical significance relative to the reference method (Hungarian-Efficiency) is denoted by $^{ns}$ (not significant) or $^{***}$ ($p < 0.001$).}
\label{tab:taxi_dispatch_comparison}
\begin{tabular}{lcccc}
\toprule
\multirow{2}{*}{\textbf{Algorithm}} & \multicolumn{2}{c}{\textbf{50 Taxis}} & \multicolumn{2}{c}{\textbf{200 Taxis}} \\
\cmidrule(lr){2-3} \cmidrule(lr){4-5}
& Revenue & Plan time (ms) & Revenue & Plan time (ms) \\
\midrule
Greedy-Nearest        & $4{,}751 \pm 42^{***}$      & $9{,}568 \pm 218$        & $17{,}401 \pm 103^{***}$    & $34{,}214 \pm 323$ \\
Greedy-Reward         & $2{,}513 \pm 86^{***}$      & $5.47 \pm 2.39$          & $7{,}876 \pm 98^{***}$      & $13.71 \pm 2.64$ \\
Greedy-Efficiency     & $6{,}916 \pm 64^{ns}$       & $249.28 \pm 12.92$       & $23{,}696 \pm 86^{ns}$      & $731.54 \pm 18.32$ \\
Hungarian-Time        & $5{,}347 \pm 49^{***}$      & $107.88 \pm 11.17$       & $19{,}840 \pm 63^{***}$     & $296.61 \pm 12.39$ \\
Hungarian-Reward      & $5{,}559 \pm 61^{***}$      & $75.20 \pm 8.79$         & $18{,}901 \pm 123^{***}$    & $237.08 \pm 14.59$ \\
Hungarian-Efficiency  & $6{,}938 \pm 60$ (ref)      & $104.83 \pm 9.13$        & $23{,}826 \pm 91$ (ref)     & $363.23 \pm 12.68$ \\
\midrule
ALNS                  & $6{,}927 \pm 62^{ns}$       & $31{,}064 \pm 1{,}023$   & $23{,}735 \pm 93^{ns}$      & $236{,}765 \pm 5{,}821$ \\
GA                    & $6{,}909 \pm 58^{ns}$       & $1.26M \pm 0.04M$        & $23{,}718 \pm 91^{ns}$      & $4.58M \pm 0.10M$ \\
SA                    & $6{,}905 \pm 58^{ns}$       & $25{,}591 \pm 746$       & $23{,}771 \pm 88^{ns}$      & $96{,}660 \pm 1{,}466$ \\
\bottomrule
\end{tabular}
\end{table*}

Taken together, the results across both domains indicate that efficiency-based assignment --- whether via a greedy or Hungarian-style matching --- consistently achieves near-optimal reward/revenue at a fraction of the computational cost of metaheuristic search. The efficiency heuristics appear robust across fleet size and domain type, making them strong candidates for real-time dispatch systems where planning time is constrained.

\section{Discussion}

This study evaluated a range of dispatch algorithms across two operationally distinct domains --- drones and taxi fleets --- with the goal of identifying approaches that balance reward performance against computational feasibility for real-time deployment. The results converge on a clear finding: efficiency-based assignment methods achieve near-optimal revenue outcomes at planning speeds that are orders of magnitude faster than metaheuristic alternatives, and substantially better revenues than simpler greedy or time/reward-optimising heuristics. The Efficiency heuristic's consistent dominance across both synthetic drone and empirical NYC taxi environments suggests the advantage reflects a structural property of dynamic reward-maximising routing rather than an artifact of the experimental setting \cite{b13,b31}.

The consistent superiority of the Efficiency heuristics across both domains suggests that the \textit{choice of optimisation criterion} --- rather than the sophistication of the assignment mechanism itself --- is the primary driver of dispatch quality. Algorithms that optimised for raw reward (Greedy-Reward, Hungarian-Reward) or travel time (Hungarian-Time) systematically underperformed, despite using comparably complex assignment procedures. This implies that neither speed nor reward alone is a sufficient proxy for fleet-level value generation; rather, the ratio of reward to cost --- efficiency --- better captures the opportunity cost of allocating a vehicle to any given job. This finding is broadly consistent with the established principle that greedy algorithms perform well when the problem exhibits optimal substructure, but that the choice of the greedy criterion fundamentally determines solution quality \cite{b32}. In the vehicle dispatch context, where committing a vehicle to one job forecloses other opportunities, efficiency-ranked assignment naturally prioritises the highest-value matches while implicitly accounting for cost \cite{b33}.

The metaheuristic algorithms (SA, ALNS, GA) achieved revenues comparable to the best-performing heuristics, confirming that longer search does eventually approach the same quality ceiling. However, the planning times required --- ranging from thousands to millions of milliseconds at moderate fleet sizes --- disqualify these methods from any real-time dispatch context. In live operations, a planning cycle that takes several minutes to complete is not merely slow; it is actively harmful, as the environment will have changed materially by the time a solution is produced. It should be noted that we used a somewhat aggressive limit on the number of iterations that GA and ALNS would run. The fact that three very different optimization schedules yield similar results supports the view that these solutions are somewhere near optimal. From a practical deployment perspective, however, the efficiency heuristics represent a more favourable point on the cost-quality frontier.

The near-equivalence of Greedy-Efficiency and Hungarian-Efficiency is a particularly noteworthy finding. The fact that the two methods yield statistically indistinguishable revenues suggests that, under the demand and supply conditions tested, the global optimality guarantee of Hungarian matching confers limited practical benefit.

While the broad pattern of results was consistent across drones and taxis, some domain-specific differences are worth highlighting. In the taxi domain, Greedy-Nearest performed considerably better relative to the reference than in the drone setting, suggesting that proximity may be a stronger dispatch signal when vehicles are operating in a dense, road-constrained network. This is consistent with prior work showing that nearest-vehicle dispatch remains a widely deployed baseline in urban ride-hailing systems. \cite{b34}.

Several limitations of the present study warrant consideration. The experiments were conducted in simulation, and the degree to which the simulated demand patterns, vehicle dynamics, and reward structures reflect real-world operational conditions is uncertain. City models that simplify complex road maps and traffic conditions may not capture important aspects of real operational environments. In particular, the results may not generalise to scenarios involving non-stationary demand, vehicle failures, or multi-objective constraints such as fairness or emissions. Work on dynamic Hungarian matching has explored how solutions can be efficiently repaired when costs change, and such approaches may offer a practical path to combining the performance quality of Hungarian-Efficiency with greater responsiveness to real-time state changes \cite{b35}. The relative advantage of metaheuristic methods may also shift in settings where longer planning horizons are acceptable. Additionally, while this study focused on revenue as the primary outcome metric, operational systems typically optimise over multiple objectives simultaneously. The interaction between efficiency-based dispatch and secondary objectives such as driver earnings equity, service coverage, or environmental cost remains to be explored. Recent work has shown that small changes in system parameters can cause large deviations in the income distributions of drivers, suggesting that revenue-maximising dispatch algorithms may carry significant unintended distributional consequences \cite{b36}.

\section{Conclusion}

Efficiency-based assignment --- scoring candidate tasks by reward per unit time cost --- consistently identifies near-optimal fleet allocations without the computational burden of metaheuristic search. Across drone and taxi domains at multiple scales, the Efficiency heuristics establish Pareto dominance on the reward-versus-compute frontier. These findings suggest that careful objective function design is more impactful than algorithmic complexity for real-time fleet dispatch. Future work should examine generalisation to broader classes of dynamic combinatorial optimization and multi-objective settings.

\section*{Acknowledgment}


\begin{thebibliography}{00}

\bibitem{b1} ``30th Annual State of Logistics Report: What's next?,'' \emph{Gale Academic OneFile}, 2019. [Online]. Available: https://go.gale.com/ps/i.do?id=GALE\%7CA596897691

\bibitem{b2} G. B. Dantzig and J. H. Ramser, ``The Truck Dispatching Problem,'' \emph{Manage. Sci.}, vol. 6, no. 1, pp. 80--91, Oct. 1959, doi: 10.1287/mnsc.6.1.80.

\bibitem{b3} G. Dantzig, R. Fulkerson, and S. Johnson, ``Solution of a Large-Scale Traveling-Salesman Problem,'' \emph{OR}, vol. 2, no. 4, pp. 393--410, Nov. 1954.

\bibitem{b4} J. Hartmanis, ``Computers and Intractability,'' \emph{SIAM Rev.}, vol. 24, no. 1, pp. 90--91, Jan. 1982.

\bibitem{b5} N. Christofides, ``Vehicle Routing,'' \emph{The Traveling Salesman}, 1985.

\bibitem{b6} H. N. Psaraftis, M. Wen, and C. A. Kontovas, ``Dynamic vehicle routing problems: Three decades and counting,'' \emph{Networks}, vol. 67, no. 1, pp. 3--31, 2016.

\bibitem{b7} M. W. Ulmer, J. C. Goodson, D. C. Mattfeld, and B. W. Thomas, ``On modeling stochastic dynamic vehicle routing problems,'' \emph{EURO J. Transp. Logist.}, vol. 9, no. 2, p. 100008, 2020.

\bibitem{b8} P. Vansteenwegen, W. Souffriau, and D. V. Oudheusden, ``The orienteering problem: A survey,'' \emph{Eur. J. Oper. Res.}, vol. 209, no. 1, pp. 1--10, 2011.

\bibitem{b9} G. S. C. Avellar, G. A. S. Pereira, L. C. A. Pimenta, and P. Iscold, ``Multi-UAV Routing for Area Coverage and Remote Sensing with Minimum Time,'' \emph{Sensors}, vol. 15, no. 11, pp. 27783--27803, 2015.

\bibitem{b10} Y. Liu, ``An optimization-driven dynamic vehicle routing algorithm for on-demand meal delivery using drones,'' \emph{Comput. Oper. Res.}, vol. 111, pp. 1--20, 2019.

\bibitem{b11} H. Billhardt, A. Fern\'{a}ndez, S. Ossowski, J. Palanca, and J. Bajo, ``Taxi dispatching strategies with compensations,'' \emph{Expert Syst. Appl.}, vol. 122, pp. 173--182, 2019.

\bibitem{b12} J. Lv, Z. Zhao, S. Yao, and W. Lv, ``Revenue-maximizing online stable task assignment on taxi-dispatching platforms,'' \emph{Front. Comput. Sci.}, vol. 16, no. 6, p. 166208, 2022.

\bibitem{b13} M. Gendreau, G. Laporte, and R. S\'{e}guin, ``A Tabu Search Heuristic for the VRP with Stochastic Demands,'' \emph{Oper. Res.}, vol. 44, no. 3, pp. 469--477, 1996.

\bibitem{b14} T. Tsiligirides, ``Heuristic Methods Applied to Orienteering,'' \emph{J. Oper. Res. Soc.}, vol. 35, no. 9, pp. 797--809, 1984.

\bibitem{b15} ``TLC Trip Record Data,'' NYC Taxi \& Limousine Commission. [Online]. Available: https://www.nyc.gov/site/tlc/about/tlc-trip-record-data.page

\bibitem{b16} H. W. Kuhn, ``The Hungarian method for the assignment problem,'' \emph{Naval Res. Logist. Q.}, vol. 2, no. 1--2, pp. 83--97, 1955.

\bibitem{b17} J. Alonso-Mora, S. Samaranayake, A. Wallar, E. Frazzoli, and D. Rus, ``On-demand high-capacity ride-sharing via dynamic trip-vehicle assignment,'' \emph{Proc. Natl. Acad. Sci.}, vol. 114, no. 3, pp. 462--467, 2017.

\bibitem{b18} P. Virtanen \emph{et al.}, ``SciPy 1.0: fundamental algorithms for scientific computing in Python,'' \emph{Nat. Methods}, vol. 17, no. 3, pp. 261--272, 2020.

\bibitem{b19} N. A. Wouda and L. Lan, ``ALNS: a Python implementation of the adaptive large neighbourhood search metaheuristic,'' \emph{J. Open Source Softw.}, vol. 8, no. 81, p. 5028, 2023.

\bibitem{b20} D. J. Rosenkrantz, R. E. Stearns, and P. M. Lewis II, ``An Analysis of Several Heuristics for the Traveling Salesman Problem,'' \emph{SIAM J. Comput.}, vol. 6, no. 3, pp. 563--581, 1977.

\bibitem{b21} G. Laporte, ``The vehicle routing problem: An overview of exact and approximate algorithms,'' \emph{Eur. J. Oper. Res.}, vol. 59, no. 3, pp. 345--358, 1992.

\bibitem{b22} B. L. Golden, L. Levy, and R. Vohra, ``The orienteering problem,'' \emph{Naval Res. Logist.}, vol. 34, no. 3, pp. 307--318, 1987.

\bibitem{b23} B. P. Gerkey and M. J. Matari\'{c}, ``A Formal Analysis and Taxonomy of Task Allocation in Multi-Robot Systems,'' \emph{Int. J. Robot. Res.}, vol. 23, no. 9, pp. 939--954, 2004.

\bibitem{b24} J. H. Holland, \emph{Adaptation in Natural and Artificial Systems}. MIT Press, 1992.

\bibitem{b25} S. Kirkpatrick, C. D. Gelatt, and M. P. Vecchi, ``Optimization by Simulated Annealing,'' \emph{Science}, vol. 220, no. 4598, pp. 671--680, 1983.

\bibitem{b26} S. Ropke and D. Pisinger, ``An Adaptive Large Neighborhood Search Heuristic for the Pickup and Delivery Problem with Time Windows,'' \emph{Transp. Sci.}, vol. 40, no. 4, pp. 455--472, 2006.

\bibitem{b27} D. Connolly, ``Knapsack Problems: Algorithms and Computer Implementations,'' \emph{J. Oper. Res. Soc.}, vol. 42, no. 6, pp. 513--513, 1991.

\bibitem{b28} W. Dinkelbach, ``On Nonlinear Fractional Programming,'' \emph{Manage. Sci.}, vol. 13, no. 7, pp. 492--498, 1967.

\bibitem{b29} A. Blum \emph{et al.}, ``Approximation Algorithms for Orienteering and Discounted-Reward TSP,'' \emph{SIAM J. Comput.}, vol. 37, no. 2, pp. 653--670, 2007.

\bibitem{b30} F. Betti Sorbelli, F. Cor\`{o}, S. K. Das, L. Palazzetti, and C. M. Pinotti, ``Drone-based Optimal and Heuristic Orienteering Algorithms Towards Bug Detection in Orchards,'' in \emph{Proc. DCOSS}, 2022, pp. 117--124.

\bibitem{b31} U. Ritzinger, J. Puchinger, and R. F. Hartl, ``A survey on dynamic and stochastic vehicle routing problems,'' \emph{Int. J. Prod. Res.}, vol. 54, no. 1, pp. 215--231, 2016.

\bibitem{b32} T. H. Cormen, C. E. Leiserson, R. L. Rivest, and C. Stein, \emph{Introduction to Algorithms}, 3rd ed. MIT Press, 2009.

\bibitem{b33} Y. Kim and Y. Yoon, ``Zone-Agnostic Greedy Taxi Dispatch Algorithm Based on Contextual Matching Matrix,'' \emph{Electronics}, vol. 10, no. 21, p. 2653, 2021.

\bibitem{b34} Y. Tong, J. She, B. Ding, L. Chen, T. Wo, and K. Xu, ``Online Minimum Matching in Real-Time Spatial Data,'' \emph{Proc. VLDB Endow.}

\bibitem{b35} G. A. Mills-Tettey, A. Stentz, and M. B. Dias, ``The Dynamic Hungarian Algorithm for the Assignment Problem with Changing Costs.''

\bibitem{b36} E. Bok\'{a}nyi and A. Hann\'{a}k, ``Understanding Inequalities in Ride-Hailing Services Through Simulations,'' \emph{Sci. Rep.}, vol. 10, no. 1, p. 6500, 2020.

\end{thebibliography}
\end{document}